\documentclass[conference]{IEEEtran}
\IEEEoverridecommandlockouts
\usepackage{amsmath,amssymb,amsfonts,bm}
\usepackage{graphicx}
\usepackage[justification=centering]{caption}
\newcommand{\sg}{\mbox{sgn}}
\usepackage[subrefformat=parens]{subcaption}
\usepackage{comment}
\def\BibTeX{{\rm B\kern-.05em{\sc i\kern-.025em b}\kern-.08em
    T\kern-.1667em\lower.7ex\hbox{E}\kern-.125emX}}
\begin{document}

\title{A Recurrent Neural Network based Clustering Method for Binary Data Sets in Education
}

\author{\IEEEauthorblockN{Mizuki Ohira}
\IEEEauthorblockA{\textit{Electrical and Electronics Engineering Dept.} \\
\textit{Hosei University}\\
Koganei, Tokyo, 184-8584, Japan\\
mic0309a@icloud.com}
\and
\IEEEauthorblockN{Toshimichi Saito}
\IEEEauthorblockA{\textit{Electrical and Electronics Engineering Dept.} \\
\textit{Hosei University}\\
Koganei, Tokyo, 184-8584, Japan\\
tsaito@hosei.ac.jp}
}
\maketitle

\begin{abstract}
This paper studies an application of a recurrent neural network to clustering method for  the S-P chart: a binary data set used widely in education. 
As the number of students increases, the S-P chart becomes hard to handle. 
In order to classify the large chart into smaller charts, we present a simple clustering method based on the network dynamics. 
In the method, the network has multiple fixed points and basins of attraction give clusters corresponding to small S-P charts. 
In order to evaluate the clustering performance, we present an important feature quantity: average caution index that characterizes singularity of students answer oatterns. 
Performing fundamental experiments, effectiveness of the method is confirmed. 
\end{abstract}
\begin{IEEEkeywords}
Recurrent neural networks, clustering, S-P chart, stability
\end{IEEEkeywords}

\section{Introduction}
This paper studies an application of a discrete-time recurrent neural network \cite{hopf} with a signum-type activation function. 
The network dynamics is described by an asynchronous autonomous difference equation of binary state variables. 
Depending on the connection parameters, the network exhibits interesting nonlinear phenomena: fixed points of binary vectors, periodic orbits of binary vectors, and related bifurcation. 
The stable fixed points can coexist and the network exhibits one of them depending on the initial condition \cite{liu}. 
Applications are many \cite{am}-\cite{kiyohara}, including 
associative memories, 
error correction codes, and reservoir computing. 

As an interesting application, we consider a clustering method of the S-P chart: a set of binary score vectors of students \cite{sp1}-\cite{sp3}. 
The S-P chart is widely used in education. 
However, as the number of students increases, the chart becomes large and becomes difficult to handle. 
In order to classify the large S-P chart into small S-P charts, we give a simple clustering method based on recurrent neural networks. 
In the method, multiple fixed points represent clusters and basins of attraction correspond to clustering. 
In order to evaluate the clustering, we define two feature quantities. 
The first quantity characterizes the distribution of the cluster size. 
The second quantity is the average caution index that characterizes the singularity of the student's response patterns. 
This second quantity is novel and important. 

Performing fundamental numerical experiments, effectiveness of the method is confirmed: an artificial large S-P chart is classified into small S-P charts and the clustering is evaluated by the feature quantities. 
Note that our method is based on the network nonlinear dynamics whereas usual clustering is based on distance between data.

\section{Preliminary}

As a preliminary, we introduce two fundamental concepts necessary for the proposed clustering method. 

\subsection{Discrete-time recurrent neural networks}
We introduce the recurrent neural networks (RNNs) consisting of $N$ signum-type neurons \cite{hopf}. 
The dynamics is described by the asynchronous autonomous difference equation:
\begin{equation}
    x_j(t+1) = 
    \left\{\begin{array}{ll}
    \displaystyle \sg \left(\sum_{j=1}^{N} w_{ij} x_j(t) \right) & \mbox{for } t = nN+j\\
    x_j(t)  & \mbox{for } t \ne nN+j
    \end{array}\right. 
    \label{dt-rnn}
\end{equation}
\[
\sg(X) = \left\{\begin{array}{ll}
+1 & \mbox{ for }  X \ge 0\\ 
-1 & \mbox{ for }  X < 0\\ 
\end{array}\right.
\]
where $x_j(t) \in \{-1, +1\}$, $j \in \{1, \cdots, N \}$, is the $j$-th bipolar state variable at discrete time $t$ and $n$ denotes positive integers.  
$w_{ij}$ denotes real valued connection parameters and is summarized into the connection matrix
\[
\bm{W} \equiv
\begin{pmatrix}
w_{11}  & \cdots  & w_{1N} \\
 \vdots & \ddots  & \vdots \\
w_{N1}  & \cdots  & w_{NN}
\end{pmatrix}
\]
$x_1(t)$ is updated at $t=1$, $x_2(t)$ is updated at $t=2$, and $x_N(t)$ is updated at time $N$. 
The update is repeated with period $T=N$ and update from $x_1(t)$ to $x_N(t)$ is completed at every $T$ iteration.  
Eq. (\ref{dt-rnn}) is abbreviated by the following vector form for every $N$ iteration:
\begin{equation}
\begin{array}{l}
\bm{x}(n+1) = F(\bm{x}(n) \bm{W}^\top )\\
 \ \bm{x}(n) \equiv (x_1(nT), \cdots, x_N(nT))
\end{array}
\end{equation}
Our clustering method is based on multiple fixed points defined as the following. 
A binary vector $\bm{p} =(p_1, \cdots, p_N)$ is said to be a fixed point 
if $\bm{p} = F(\bm{p})$. 
A set of binary vectors $A=\{ \bm{a}_1, \cdots, \bm{a}_{L_a} \}$ is said to be 
a basin of attraction of the fixed point $\bm{p}$ if 
$F^l(\bm{a}_i) = \bm{p}$, $\forall \bm{a}_i \in A$, $i \in \{1, \cdots, L_a\}$ 
where $F^l$ is the $l$-fold composition of $F$. 
A sequence started from an initial point falls into some fixed point if the following condition is satisfied \cite{hopf}: 
\begin{equation}
w_{ij} = w_{ji} \ (i \ne j), \ w_{ii}=0.
\label{con1}
\end{equation}

\subsection{S-P charts}

We introduce the S-P chart (Student - Problem chart): a useful data set consisting of binary score vectors of students. 
Extracting simple feature quantities, the S-P chart provides helpful information for teachers to instruct students \cite{sp1}. 
Let $L$ students answer a test consisting of $N$ problems. 
Let $d_{ij}$ denote answer of the $i$-th student on the $j$-th problem 
where $d_{ij}=0$ indicates an incorrect answer and $d_{ij}=1$ indicates a correct answer. 
The result of the $i$-th student is summarized into a binary score vector
\begin{equation}
\bm{d}_i \equiv (d_{i1}, \cdots, d_{iN}), \; i \in \{1, \cdots, L \}, \; d_{ij} \in \{0, 1\} 
\label{bcv}
\end{equation}
The results of $L$ students are summarized into an $L \times N$ binary matrix as shown in Fig. \ref{SP_Overview} (a): 
\begin{equation}
D \equiv 
\begin{pmatrix}
d_{11}  & \cdots  & d_{LN} \\
 \vdots & \ddots  & \vdots \\
d_{L1}  & \cdots  & d_{LN}
\end{pmatrix}
\label{dset}
\end{equation}
We refer to this matrix as a primitive S-P chart.  
It is equivalent to a binary data set of $L$ binary score vectors. 
For convenience, 
in the rows, the order of the student scores is rearranged from high to low.   
In the columns, the order of the problems is rearranged from easy to difficult:  
\begin{equation}
\begin{array}{l}
\displaystyle S(i) \equiv \sum_{j=1}^N d_{ij}, \  S(1) \ge \cdots \ge S(L)\\
\displaystyle P(j) \equiv \sum_{i=1}^L d_{ij}, \  P(1) \ge \cdots \ge P(N)
\end{array}
\label{spsum}
\end{equation}
In the rearranged chart, we mark two curves (see Fig. \ref{SP_Overview} (b)): 
\[
\begin{array}{ll}
\mbox{S-curve: }  (i, S(i)) & i \in \{1, \cdots, L \}\\
\mbox{P-curve: }  (P(j), j) & j \in \{1, \cdots, N \}
\end{array}
\]
The S-curve represents the distribution of student scores. 
The P-curve represents the distribution of correct answers for problems. 
The rearranged matrix with the two curves is referred to as the S-P chart. 
Fig. \ref{SP_Typical} shows three fundamental types of the S-P chart:
\begin{figure}[htb]
    \centering
    \includegraphics[width=0.8\columnwidth]{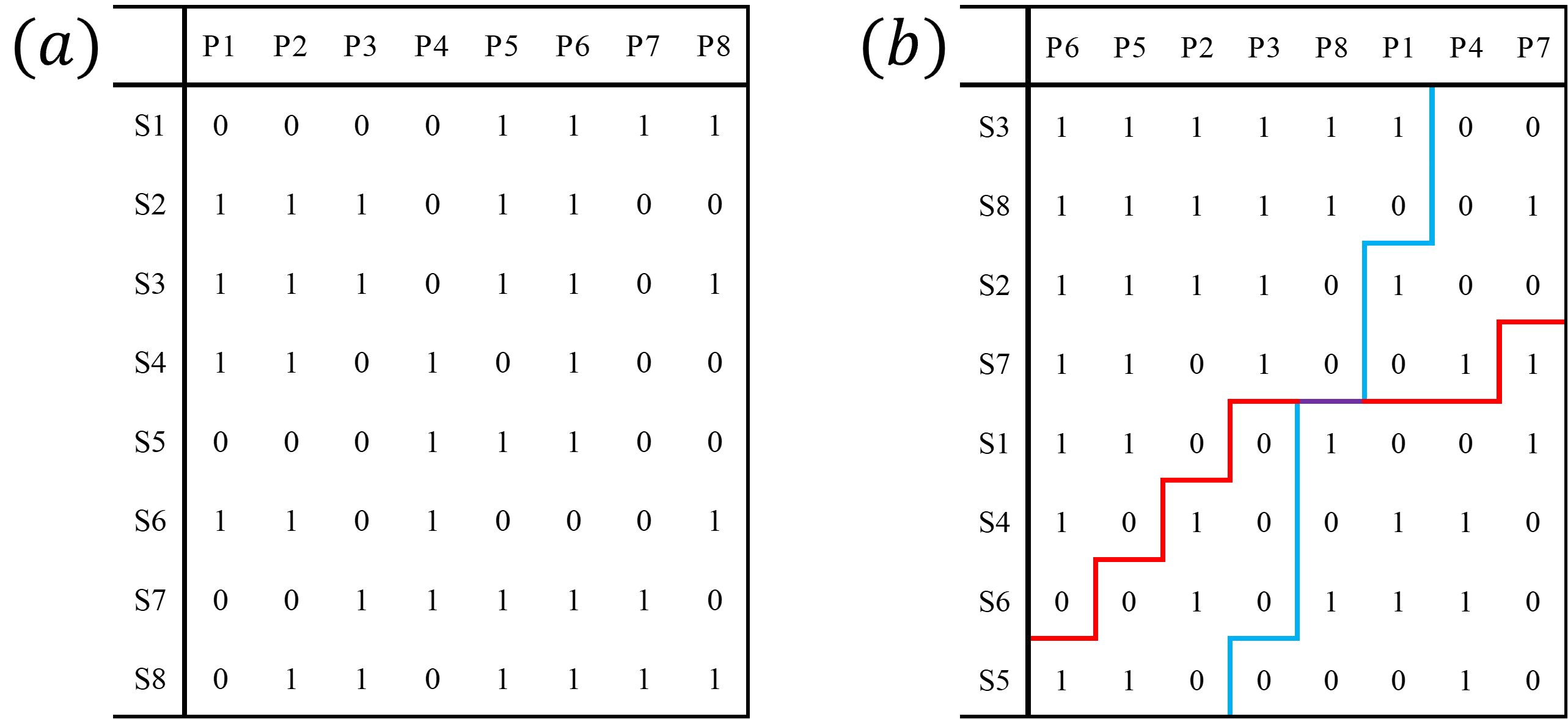}
\caption{S-P chart. 
(a) Primitive S-P chart of binary score vectors. 
(b) S-P chart with S- and P-curves. }
    \label{SP_Overview} 

\vspace*{3mm}

    \begin{tabular}{ccc}
      \begin{minipage}[t]{0.32\columnwidth}
        \centering
        \includegraphics[width=0.8\columnwidth]{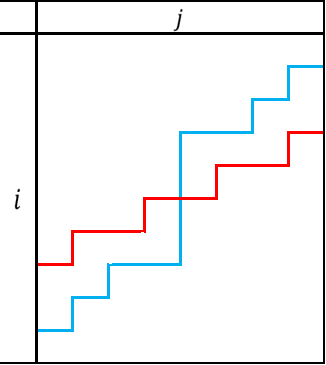}
        \subcaption{Test type.}
      \end{minipage} 
      \begin{minipage}[t]{0.32\columnwidth}
        \centering
        \includegraphics[width=0.8\columnwidth]{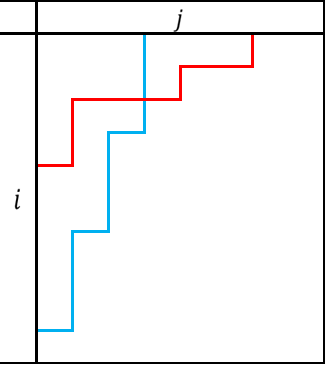}
        \subcaption{Pre-test type.}
      \end{minipage} 
      \begin{minipage}[t]{0.32\columnwidth}
        \centering
        \includegraphics[width=0.8\columnwidth]{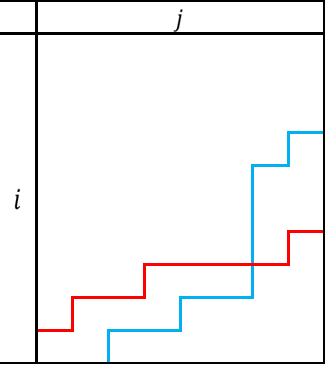}
        \subcaption{Drill type.}
      \end{minipage} 
    \end{tabular}
     \caption{Fundamental three types of the S-P chart.}
     \label{SP_Typical}
  \end{figure}

\begin{itemize}

\item Test type: 
The correct answer rate for each problem is about 50\%. 
The student score distribution is almost normal distribution. 

\item Drill type: 
The correct answer rate for each problem is about 65\% or more (easy problems). 
The number of high-score students is large. 

\item Pre-test type:
The correct answer rate for each problem is about 35\% or less (hard problems).  
the number of low-score students is large. 

\end{itemize}

\section{Clustering of large S-P charts}

As the number of students ($L$) increases, the S-P chart becomes large and becomes difficult to handle. 
For convenience in the analysis, we present a clustering method corresponding to classification of students: the large S-P chart is classified into small suitable S-P charts. 
In order to formulate the method, we give several notations. 
Let the matrix $D$ of Eq. (\ref{dset}) be a large S-P chart after the rearrangement. 
The chart $D$ is equivalent to a set of binary score vectors:  
\begin{equation}
D = \{\bm{d}_1, \cdots, \bm{d}_L \}, \ 
\bm{d}_i \equiv (d_{i1}, \cdots, d_{iN}), \ 
d_{ij} \in \{0, 1\}
\end{equation}
where $i \in \{1, \cdots, L \}$. 
Let $C$ be a set of small S-P charts and 
let $D^l$ be the $l$-th small S-P chart consisting of $L_l$ binary score vectors $\bm{d}^l_i$ selected from $D$: 
\begin{equation}
\begin{array}{l}
C = \{D^1, \cdots, D^{M'} \}\\
D = D^1 \cup  \cdots \cup D^{M'}, \; 
D^l \cap D^k = \emptyset, \; l \neq k\\ 
D^l = \{d^l_1, \cdots, d^l_{L_l} \}, \; 
\bm{d}^l_i \equiv (d^l_{i1}, \cdots, d^l_{iN}), \; 
d^l_{ij} \in \{0, 1\}
\end{array}
\end{equation}
The small S-P charts are convenient for analysis/investigation. 
It provides helpful information for student instruction.

\subsection{Feature quantities}

In order to evaluate the clustering performance, 
we introduce two feature quantities of a clustering $C$. 
The first quantity is defined by
\begin{equation}
f_1(C) = \frac{L_D - L_M(C)}{L_D}, \qquad  L_D = L/M
\end{equation}
where $M$ is the desired number of clusters, 
$L_D$ is the desired cluster size, 
and $L_M(C)$ is the size of the $M$-th largest cluster.
$L$ is the total number of students. 
As $f_1$ decreases, the size of clusters (the number of students in each cluster) approaches to  uniform distribution.
If $f_1=0$ then all clusters have the same number of students.

As the major/novel contribution of this paper, we present the second quantity: 
\begin{equation}
\begin{array}{l}
\displaystyle f_2(C) = \max_l \gamma^l, \qquad \gamma^l = \frac{1}{L_l}\sum_{i=1}^{L_l} \gamma^l_i \\

\displaystyle \mbox{Caution index}: \ \gamma^l_i = \frac{1}{N} \sum_{j=1}^{N} | d^l_{ij} - \mu^l_j |
\end{array}
\end{equation}
where $\gamma^l_i$ is the caution index of the $i$-th student in the $l$-th cluster and $\mu^l_j$ is the correct answer rate of the $j$-th problem in the $l$-th cluster. 
The caution index $\gamma^l_i$ characterizes the answer singularity of a student in a cluster. 
$\gamma^l$ is the average of the caution index for students in the $l$-th cluster. 
The average caution index $\gamma^l$ depends on the clustering. 
As the $\gamma^l$ decreases, the number of answer singularity decreases in each cluster. 
Replacing $\mu^l_j$ and $\bm{d}^l_i$ with 
correct answer rate $\mu_j$ and binary score vector $\bm{d}_i$ in the larger S-P chart, respectively, we obtain the average caution index of the large S-P chart.

\subsection{DT-RNN based Clustering algorithm}
The clustering algorithm is defined by the following 4 steps. \\

\noindent
{\bf STEP 1: }
Randomly select $M$ representative
binary score vectors $\bm{r}_i$ from $D$:
\[
\bm{r}_l\in{D}, \: l\in\{1, \cdots, M\}, \;
\bm{r}_l \equiv\{r_{l1}, \cdots, r_{lN}\} 
\]

\noindent
{\bf STEP 2: }
Applying the correlation-based learning to the representative score vectors, we obtain the connection parameters:  
\begin{equation}
   w_{ij} = \sum_{l=1}^M (2r_{li}-1)(2r_{lj}-1), \quad
   w_{ii} = 0
\label{wij}
\end{equation}
Since Eq. (\ref{wij}) satisfies the condition of Eq. (\ref{con1}), a solution started from any initial condition must fall into some fixed point. 
Let the DT-RNN have $M'$ fixed points
\footnote{
The number of fixed points depends on selection of the representative score vectors. 
If the selection is suitable, the DT-RNN has $M$ fixed points. 
}.\\[3mm] 
\noindent
{\bf STEP 3: }
For $i=1, \cdots, L$, do. \\
Apply the $i$-th score vector $\bm{d}_i$ to initial value of the DT-RNN. 
\[
\bm{x}(0)=(x_1(0), \cdots, x_N(0)), \ x_j(0) = 2d_{ij} - 1
\]
If the solution $\bm{x}(n)$ falls into the $k$-th fixed point, 
then $\bm{d}_i$ is classified into the $k$-th cluster (small S-P chart)$D_k$. 
\begin{equation}
    \bm{d}_i \in D_l \quad \text{if} \quad \bm{p}_l = F(\bm{d}_l)
\end{equation}
The number of clusters is at most $M'$. \\[3mm]
\noindent
{\bf STEP 4: }
The clusters $C$ are evaluated by the cost functions $f_1(C)$ and $f_2(C)$.

\section{Numerical experiments}
\begin{figure}[t!]
  \centering
  \includegraphics[width=0.9\linewidth]{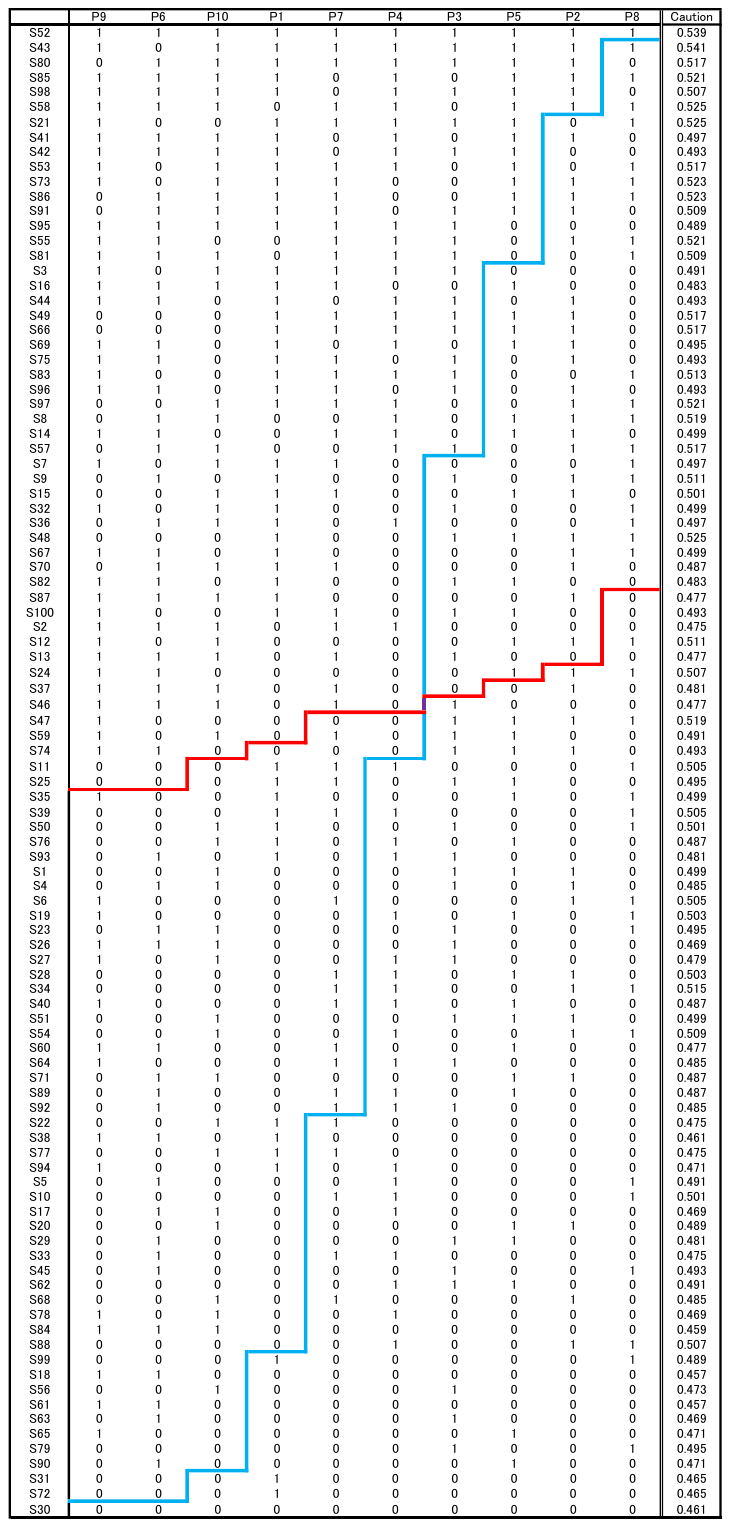}
  \caption{Large S-P Chart (Dataset $D$)}
  \label{lsp}
\end{figure}

We make an artificial large S-P chart as shown in Fig. \ref{lsp}: 

 \ \ The number of students $L=100$, 

 \ \ The number of problems $N=10$, 

 \ \ Average Caution index $\gamma = 0.494$.

\begin{figure}[htb]
  \centering
  \includegraphics[width=1\columnwidth]{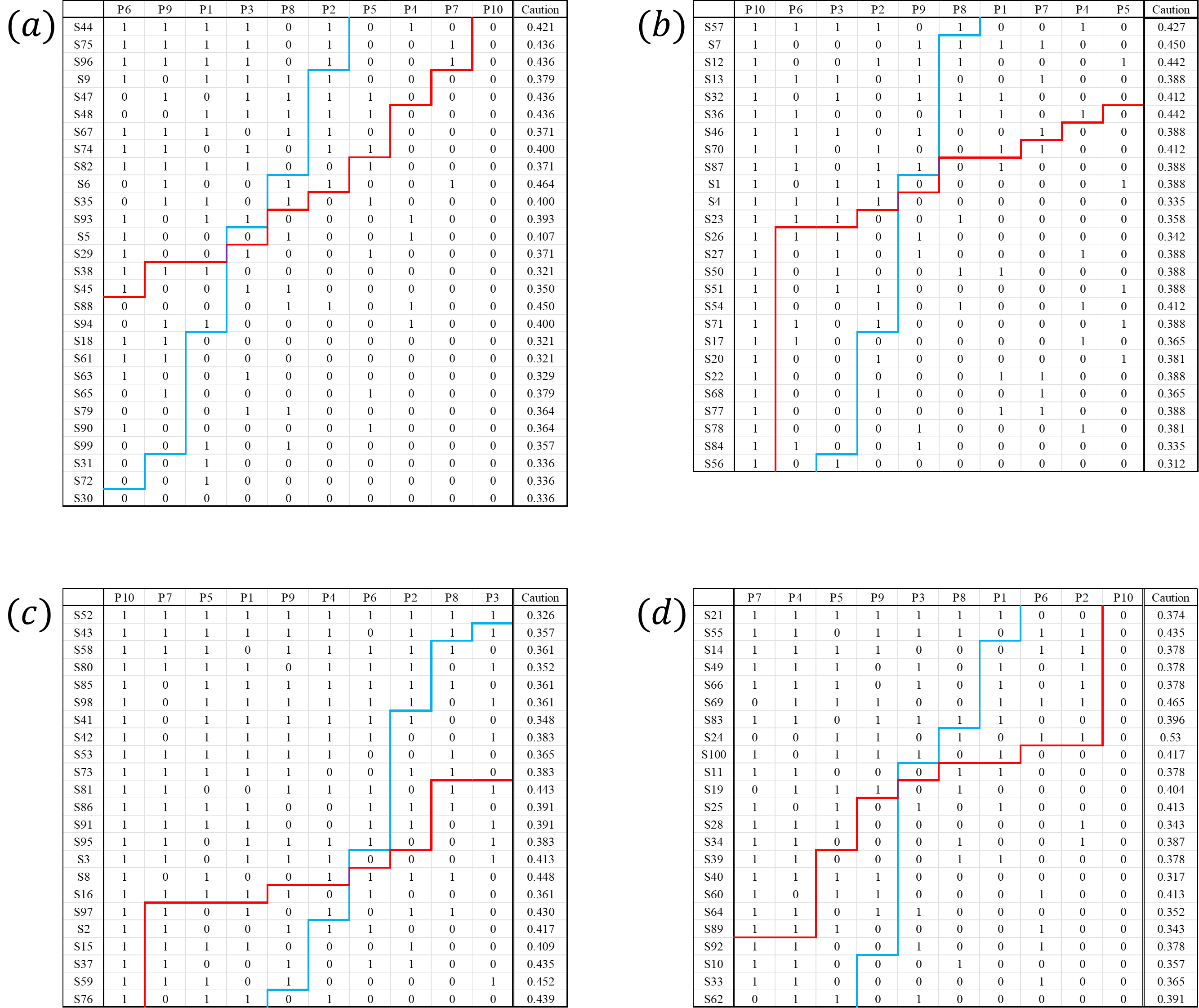}
  \caption{Small S-P Charts. \\
  (a) $C_1$, Pre-test type.  
  (b) $C_2$, Pre-test type. \\
  (c) $C_3$, Drill type. 
  (d) $C_4$, Pre-test type. 
  }
  \label{ssp}
\end{figure}

\noindent
We have tried to classify this large S-P chart into small S-P charts with lower average caution indices.
Here, we show a typical clustering result in
10,000 trials. 
In Step 1, we select the $M=4$ representative binary score vectors: 
\begin{equation}
    \begin{array}{l}
        \mbox{S32}~:~\bm{r}_1 = (1, 0, 1, 0, 0, 0, 0, 1, 1, 1)\\
        \mbox{S40}~:~\bm{r}_2 = (0, 0, 0, 1, 1, 0, 1, 0, 1, 0)\\
        \mbox{S58}~:~\bm{r}_3 = (0, 1, 0, 1, 1, 1, 1, 1, 1, 1)\\
        \mbox{S63}~:~\bm{r}_4 = (0, 0, 1, 0, 0, 1, 0, 0, 0, 0)
    \end{array}
  \label{R}
\end{equation}
In Step 2, we obtain the following correlation matrix $W$ that gives 4 fixed points $\bm{p}_1$ to $\bm{p}_4$: 
{\small
\begin{equation}
    W=
    \begin{pmatrix}
0 & 0 & 2 & -2 & 2 & -2 & -2 & -2 & 0 & 2\\
0 & 0 & -2 & 2 & 2 & 2 & 2 & 2 & 0 & 2\\
2 & -2 & 0 & -4 & -4 & 0 & -4 & 0 & -2 & 0\\
-2 & 2 & -4 & 0 & 4 & 0 & 4 & 0 & 2 & 0\\
-2 & 2 & -4 & 4 & 0 & 0 & 4 & 0 & 2 & 0\\
-2 & 2 & 0 & 0 & 0 & 0 & 0 & 0 & -2 & 0\\
-2 & 2 & -4 & 4 & 4 & 0 & 0 & 0 & 2 & 0\\
 2 & 2 & 0 & 0 & 0 & 0 & 0 & 0 & 2 & 4\\
 0 & 0 & -2 & 2 & 2 & -2 & 2 & 2 & 0 & 2\\
 2 & 2 & 0 & 0 & 0 & 0 & 0 & 4 & 2 & 0\\
    \end{pmatrix}
    \label{W}
\end{equation}
}
\begin{equation}
    \begin{array}{l}
        \bm{p}_1 = (1, 0, 1, 0, 0, 0, 0, 0, 0, 0)\\
        \bm{p}_2 = (1, 0, 1, 0, 0, 0, 0, 1, 0, 1)\\
        \bm{p}_3 = (0, 1, 0, 1, 1, 1, 1, 1, 1, 1)\\
        \bm{p}_4 = (0, 1, 0, 1, 1, 1, 1, 0, 1, 0)
    \end{array}
  \label{P}
\end{equation}
In Step 3. we obtain 4 clusters $C_1$ to $C_4 $ corresponding to 4 small S-P charts as shown in Fig. \ref{ssp}. 
Using the feature quantities, this clustering is evaluated as shown in Table \ref{tb1}. 
We can see that the clustering size is almost uniform and caution indices are lower than the large S-P chart: the students classification is convenient in instruction. 
Such a classification is difficult in the score-based clustering as shown in Table \ref{tb2}. 
It goes without saying that the student distribution is uniform, however, the caution indices are higher than the DT-RNN based clustering. 

\begin{table}[htb]
    \centering
\begingroup
\renewcommand{\arraystretch}{1.2}
    \caption{DT-RNN based clustering\\
    ($f_1(C)=0.080, \; f_2(C)=0.392$)}
    \label{tb1}
    \begin{tabular}{|c|c|c|c|c|c|c|} \hline
Cluster & $C_1$ & $C_2$ & $C_3$ & $C_4$ \\
    \hline
\# students  $L_l$ & 28 & 26 & 23 & 23 \\
    \hline
Caution $\gamma^l$ & 0.382 & 0.387 & 0.392 & 0.390 \\
    \hline
    \end{tabular}
    \label{RNN Cluster}
\endgroup
\end{table}

\begin{table}[htb]
    \centering
\begingroup
\renewcommand{\arraystretch}{1.2}
    \caption{Score based Clustering\\
    ($f_1(C)=0.000, \; f_2(C) = 0.458$)}
    \label{tb2}
    \begin{tabular}{|c|c|c|c|c|c|c|} \hline
Cluster & $C_1$ & $C_2$ & $C_3$ & $C_4$ \\
    \hline
\# Students  $L_l$ & 25 & 25 & 25 & 25 \\
    \hline
Caution $\gamma^l$  & 0.404 & 0.454 & 0.458 & 0.348 \\
    \hline
    \end{tabular}
    \label{Score Cluster}
\endgroup
\end{table}

\clearpage

\section{Conclusions}
Application of the DT-RNN to the clustering of a large S-P chart is considered in this paper. 
The DT-RNN has multiple fixed points whose basin of attraction gives a cluster corresponding to a small S-P chart.  
In order to evaluate the clustering performance, we present a novel feature quantity: the average caution index that characterizes the singularity of students' answers. 
Performing fundamental numerical experiments, it is confirmed that our method can classify a larger S-P chart into smaller S-P charts with lower caution indices. 

Future problems include the following:
automatic selection algorithm of representative score vectors for effective clustering, 
application to various types of large S-P charts, and 
application to S-P charts in real education environments.


\end{document}